\documentclass[letterpaper,journal]{IEEEtran}

\usepackage{amsmath,amsfonts,amssymb}
\usepackage{graphicx}
\usepackage{booktabs}
\usepackage{multirow}
\usepackage{makecell}
\usepackage{xcolor}
\usepackage{colortbl}
\usepackage[caption=false,font=footnotesize]{subfig}
\usepackage{textcomp}
\usepackage{stfloats}
\usepackage{url}
\usepackage{cite}
\usepackage{caption,subcaption}
\usepackage[hidelinks]{hyperref}

\begin{document}
	
	\title{Rethinking Exposure Correction for Spatially Non-uniform Degradation}
	
	\author{Ao Li, Jiawei Sun, Le Dong $^\dagger$, Zhenyu Wang, Mingtao Feng, Weisheng Dong $^\dagger$,~\IEEEmembership{Member, IEEE} and Xin Li,~\IEEEmembership{Fellow, IEEE}
		\thanks{$^\dagger$Le Dong and Weisheng Dong are the corresponding authors.
			
			This work was supported in part by the Natural Science Foundation of China under Grant 62472336, Grant 62302369 and Grant U2570205, and the Fundamental Research Funds for the Central Universities under Grant QTZX23005.}
		
		\thanks{Ao Li, Jiawei Sun, Le Dong, Mingtao Feng and Weisheng Dong are with the School of Artificial Intelligence, Xidian University, Xi'an 710071, China (e-mails: ali\_0607@stu.xidian.edu.cn;23009200867@stu.xidian.edu.cn; dongle@xidian.edu.cn; mtfeng@xidian.edu.cn; wsdong@mail.xidian.edu.cn).}
		\thanks{Zhenyu Wang is with the Hangzhou Institute of Technology, Xidian University, Hangzhou 311231, China.(e-mails: wangzhenyu@xidian.edu.cn)}
		\thanks{Xin Li is with the Department of Computer Science, University at Albany, State University of New York, Albany, NY 12222, USA (e-mail: xli48@albany.edu).}
	}
	
	\markboth{Journal of \LaTeX\ Class Files,~Vol.~14, No.~8, August~2021}%
	{Shell \MakeLowercase{\textit{et al.}}: A Sample Article Using IEEEtran.cls for IEEE Journals}
	
	\IEEEpubid{0000--0000/00\$00.00~\copyright~2021 IEEE}

	\maketitle
	
	\begin{abstract}
		Real-world exposure correction is fundamentally challenged by spatially non-uniform degradations, where diverse exposure errors frequently coexist within a single image. However, existing exposure correction methods are still largely developed under a predominantly uniform assumption. Architecturally, they typically rely on globally aggregated modulation signals that capture only the overall exposure trend. From the optimization perspective, conventional reconstruction losses are usually derived under a shared global scale, thus overlooking the spatially varying correction demands across regions. To address these limitations, we propose a new exposure correction paradigm explicitly designed for spatial non-uniformity. Specifically, we introduce a Spatial Signal Encoder to predict spatially adaptive modulation weights, which are used to guide multiple look-up tables for image transformation, together with an HSL-based compensation module for improved color fidelity. Beyond the architectural design, we propose an uncertainty-inspired non-uniform loss that dynamically allocates the optimization focus based on local restoration uncertainties, better matching the heterogeneous nature of real-world exposure errors. Extensive experiments demonstrate that our method achieves superior qualitative and quantitative performance compared with state-of-the-art methods. Code and pre-trained models are available at \url{https://github.com/FALALAS/rethinkingEC}.
	\end{abstract}
	
	
	\section{Introduction}
	
\begin{figure}[!t]
		\centering
		\subfloat[Camera live-view snapshots: exposure errors are predominantly global in low-light scenes but spatially non-uniform in high-dynamic-range scenes.\label{fig:reason}]{%
			\includegraphics[width=\linewidth]{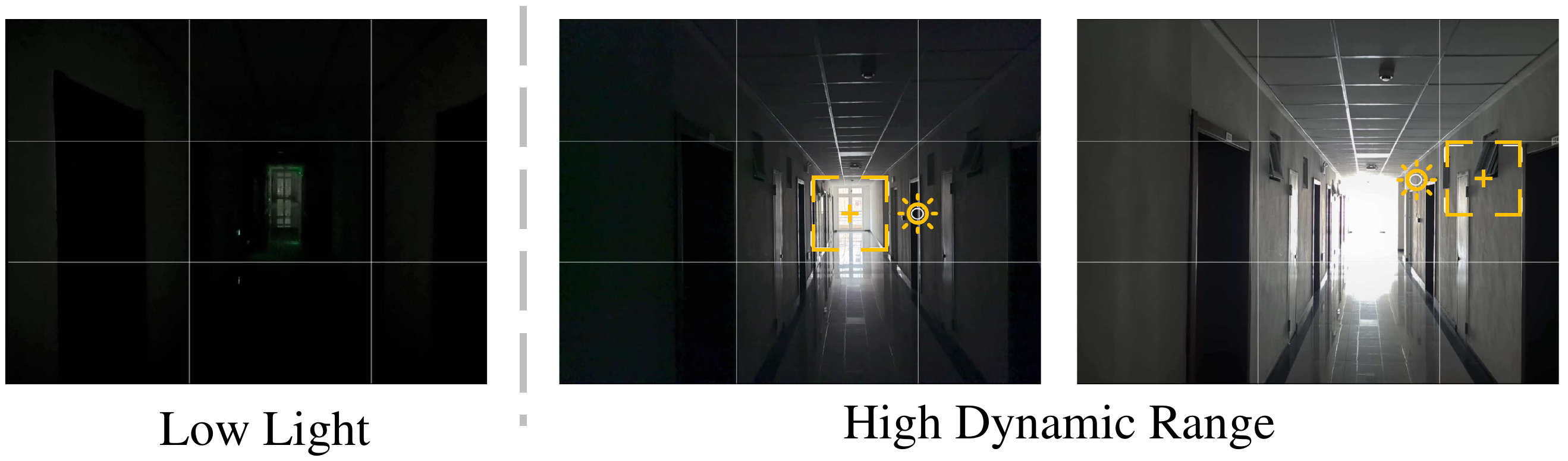}}
		\\[2pt]
		\subfloat[Visual comparison with CoTF~\cite{cotf} and LACT~\cite{lact}. Even state-of-the-art methods exhibit globally biased correction, whereas ours faithfully recovers the scene dynamic range.\label{fig:titlecom}]{%
			\includegraphics[width=\linewidth]{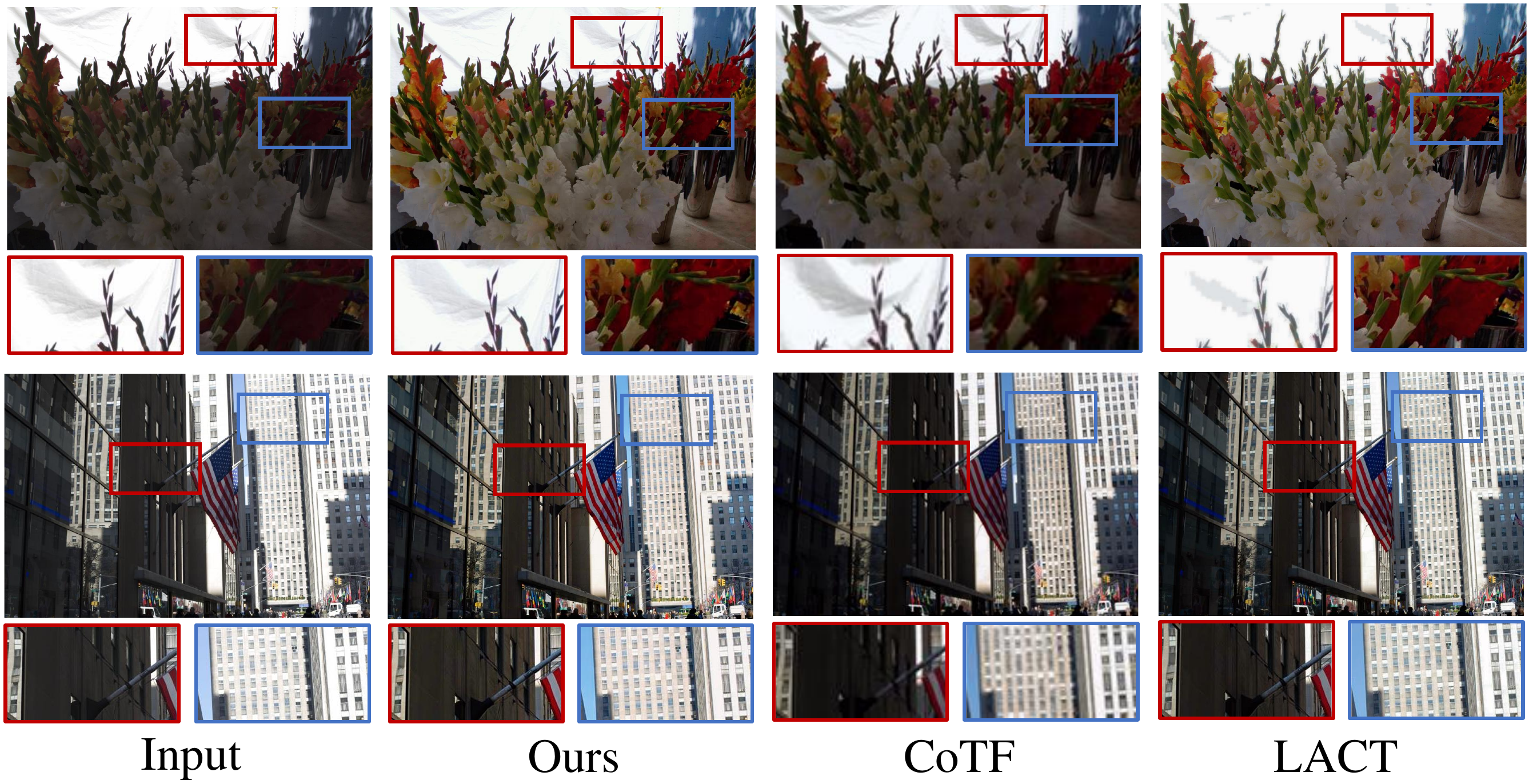}}
		\caption{Motivation for our spatially non-uniform exposure correction.}
		\label{fig:moti}
	\end{figure}

	\IEEEPARstart{E}{xposure} correction aims to restore visually faithful images from incorrectly exposed inputs. In real-world photography, exposure errors mainly arise in two representative scenarios, namely, low-light scenes and high-dynamic-range (HDR) scenes. In low-light conditions, images primarily suffer from overall under-exposure, where restoration largely relies on global brightening. Meanwhile, in HDR scenes, the wide illumination span exceeds the limited dynamic range of camera sensors. Even with modern auto-exposure systems, properly exposing the metered region often comes at the expense of losing details in brighter or darker areas, as shown in Fig.~\ref{fig:reason}. Consequently, real-world exposure errors extend beyond uniform brightness shifts, establishing spatially non-uniform degradation as an intrinsic characteristic of exposure correction.

	\IEEEpubidadjcol
	
	Early efforts in a closely related subtask, low-light image enhancement~\cite{zerodce++,pairlie,ruas,sci,hvi,sid,zerodce,retinexnet}, primarily target dark environments dominated by under-exposure, where restoration centers on uniform luminance recovery. In recent years, research attention has increasingly shifted toward general exposure correction, which seeks to handle both under- and over-exposure within a unified framework~\cite{cotf,fecnet,mmht,reed,lact,lcdp,recnet,ecl,enc}. Despite notable progress, existing methods still exhibit globally biased correction behavior and fall short of meeting the spatially varying restoration demands of real-world scenes. While prior work such as LCDP~\cite{lcdp} moves one step closer by curating a dedicated non-uniform dataset, spatially varying degradations are in fact already prevalent in standard benchmarks. As shown in Fig.~\ref{fig:titlecom}, on challenging non-uniform samples from the widely used MSEC dataset~\cite{msec}, recent state-of-the-art approaches such as CoTF~\cite{cotf} and LACT~\cite{lact} still suffer from noticeable global bias, whereas our method achieves substantially better restoration. This contrast suggests that the current performance bottleneck stems less from dataset availability and more from the underlying algorithmic design.
	
	We argue that the fundamental limitation of existing exposure correction methods lies in their implicit uniformity assumption, which pervades both network architectures and optimization objectives. From an architectural perspective, most methods can be viewed as following a common two-step paradigm in which a modulation signal is first estimated and then used to guide the subsequent image transformation. 
	However, whether derived from Laplacian pyramids~\cite{msec,mmht}, Fourier-domain amplitudes~\cite{fecnet,osmamba}, or encoder-extracted features~\cite{cotf,lact,lcdp,reed}, these signals are typically estimated from aggregated representations and therefore remain predominantly global. Consequently, such designs tend to impose uniform correction patterns across the image and struggle to accommodate region-dependent restoration demands, as analyzed in Sec.~\ref{sec:arch} and empirically verified in Sec.~\ref{sec:modulation_abla}. Beyond network architecture, a similar limitation persists in the optimization objective. Drawing inspiration from uncertainty modeling in low-level vision~\cite{ningqian,fangzhenxuan}, we revisit exposure correction losses from a probabilistic perspective. As derived in Sec.~\ref{sec:l1l2}, widely used reconstruction losses such as $L_1$ and $L_2$ are implicitly rooted in homoscedastic assumptions, \emph{i.e.}, all pixels are optimized under a shared global scale. Such a formulation conflicts with the spatially varying correction demands of mixed exposure conditions. Therefore, effective exposure correction should explicitly model spatial non-uniformity not only in the network architecture but also in the optimization objective.

	Based on these observations, we propose a new exposure correction paradigm that jointly models spatial non-uniformity at both the architectural and optimization levels. At the architectural level, rather than relying on globally aggregated descriptors, we leverage a Spatial Signal Encoder~\cite{restormer} to predict dense, spatially varying modulation weights, which subsequently guide multiple 3D look-up tables (LUT)s~\cite{3dlut} for precise image transformation. In addition, to alleviate color distortion caused by manipulating coupled sRGB channels, we incorporate a Hue-Saturation-Lightness (HSL)-based compensation branch. At the optimization level, we introduce an uncertainty-inspired non-uniform loss to guide spatially adaptive supervision. Unlike conventional uncertainty modeling in low-level vision that equates uncertainty with restoration difficulty, our formulation leverages dense uncertainty estimation to distinguish regions with distinct exposure degradations and dynamically balance their optimization weights.

	We evaluate the proposed method on four exposure correction benchmarks, namely MSEC~\cite{msec}, SICEV2~\cite{enc}, LCDP~\cite{lcdp}, and REED~\cite{reed}. Extensive experiments demonstrate that our method consistently achieves state-of-the-art performance in both quantitative metrics and visual quality. Beyond the standard supervised setting, we further validate our framework under unsupervised training, where it maintains a clear advantage over recent unsupervised approaches. Our main contributions are summarized as follows:

	\begin{itemize} 
		\item We attribute the core challenge of real-world exposure correction to the implicit uniformity assumption underlying existing network architectures and training objectives, rather than to data availability.
		
		\item Architecturally, we propose a spatially adaptive paradigm in which a Spatial Signal Encoder predicts dense, spatially varying modulation weights to guide 3D LUTs for image transformation, complemented by an HSL-based compensation branch for color fidelity.
		
		\item At the optimization level, we revisit conventional reconstruction losses from a probabilistic perspective and propose an uncertainty-inspired non-uniform loss that distinguishes regions with distinct exposure degradations and dynamically balances their optimization weights.
	\end{itemize}

	\section{Related Works}
	
	\subsection{Exposure Correction}
	
	Since the pioneering work MSEC~\cite{msec}, unified frameworks for correcting mixed exposure degradations have attracted increasing attention. Initial works primarily focused on modeling exposure-invariant representations \cite{enc} and applying them to existing low-light enhancement methods \cite{sid,drbn}. Along this line, several methods have adopted a dual-illumination estimation paradigm~\cite{lcdp,csec,ecl}, where the input image and its inverted counterpart are jointly processed and subsequently fused to yield the final correction. However, by emphasizing exposure-invariant information, such methods mainly capture global exposure cues and tend to suppress the local discrepancies that are essential for spatially non-uniform correction. Furthermore, other approaches have explored alternative representation spaces for illumination modulation. For instance, FECNet~\cite{fecnet} discovered that manipulating Fourier-domain amplitudes can adjust overall exposure, while MSEC~\cite{msec} achieved similar effects by swapping the top-level features of a Laplacian pyramid. More recently, LACT~\cite{lact} learns overall illumination through order learning~\cite{orderlearning}, and CoTF~\cite{cotf} uses the weights of 3D LUTs~\cite{3dlut} to represent image-level illumination. 
	
	Despite diverse formulations, these methods are still largely built upon a shared assumption that exposure can be characterized by a global or near-global modulation signal, typically represented in an aggregated feature space or even reduced to an orderable scalar. In contrast, our method explicitly models spatially varying correction demands through dense non-uniform modulation and LUT-based transformation.
	
	\begin{figure*}[t]
		\centering
		\includegraphics[width=\textwidth]{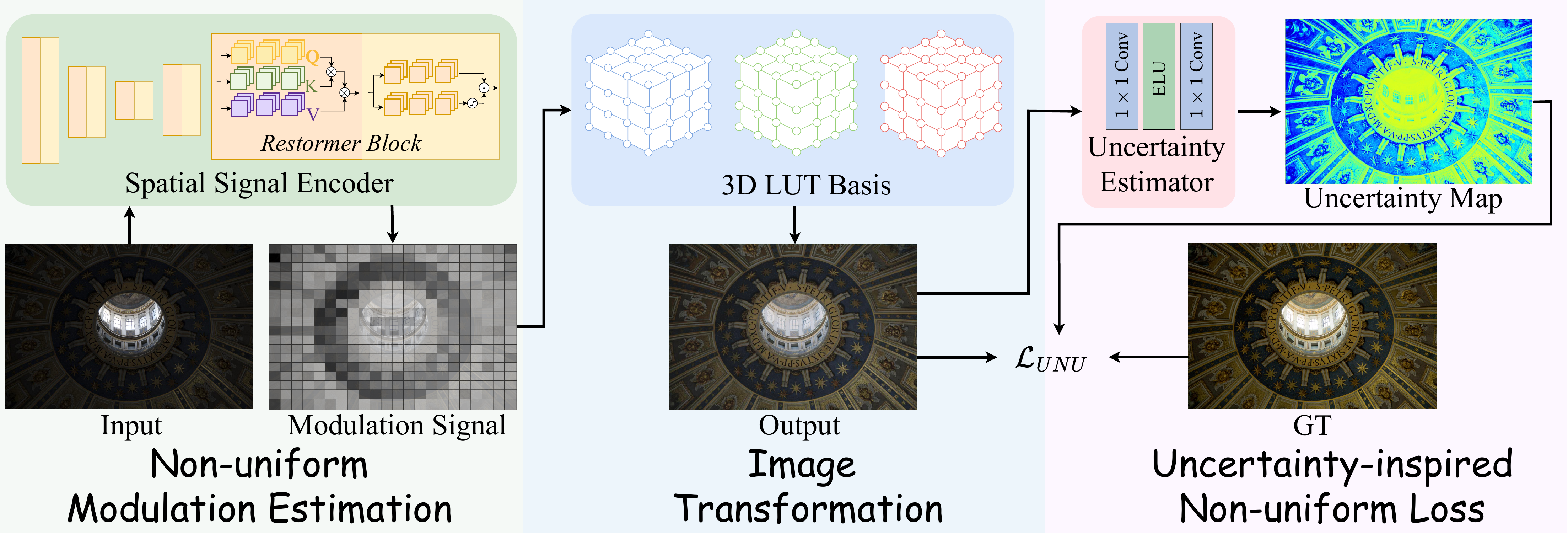}
		\caption{
			Overview of the proposed framework. 
			Given an incorrectly exposed input image, our method first performs non-uniform modulation estimation with a Spatial Signal Encoder to generate a spatially varying modulation signal. 
			The estimated signal is then used to explicitly guide multiple 3D LUT bases for precise image transformation, yielding the restored output image. 
			During training, an Uncertainty Estimator predicts a dense uncertainty map, which is used to construct the proposed uncertainty-inspired non-uniform loss $\mathcal{L}_{UNU}$ with the ground truth image. 
			In this way, our framework jointly models spatial non-uniformity at both the architectural and optimization levels.
		}
		\label{fig:framework}
	\end{figure*}
	
	\subsection{Uncertainty Modeling in Low-level Vision}
	
	In low-level vision tasks, uncertainty modeling has been widely explored to guide non-uniform model learning. For instance, uncertainty-driven mechanisms have been applied to estimate pixel-wise variance in super-resolution \cite{ningqian,fangzhenxuan}, model non-uniform transmission maps in dehazing \cite{wangbokang}, capture local heterogeneity in snow removal \cite{uncertaintyDerain,uncertaintyDesnow}, and represent non-uniform kernel priors in deblurring \cite{fangzhenxuan}. These studies collectively suggest that uncertainty modeling is highly effective for handling spatially varying degradations, making it a natural choice for exposure correction. However, previous methods typically rely on sparsity-promoting priors~\cite{jeffreysPrior}, which assume that high-uncertainty regions are sparsely distributed. In contrast, exposure degradations, such as clipped highlights and deep shadows, often appear in dense and spatially continuous forms. Therefore, instead of enforcing sparsity, our uncertainty modeling is designed to be dense and region-adaptive. More importantly, rather than serving as a conventional indicator of restoration difficulty, our estimated uncertainty acts as a dynamic spatial weight modulator. Since exposure correction involves distinct bidirectional degradations, this formulation enables the model to capture spatially varying residual distributions and better balance the heterogeneous correction demands across different image regions.

	\section{Methodology}
	
	In this section, we detail the proposed framework for exposure correction, as illustrated in Fig.~\ref{fig:framework}. First, we present a spatially adaptive architectural paradigm (Sec.~\ref{sec:arch}) that follows the two-step formulation but makes the modulation estimation spatially non-uniform, guiding image transformation through 3D LUTs to effectively handle mixed exposures. Second, we introduce an uncertainty-inspired non-uniform loss (Sec.~\ref{sec:l1l2}) to dynamically balance the learning focus across regions with different degradation directions.
	
	\subsection{Spatially Adaptive Architecture for Non-uniform Exposure Correction}
	\label{sec:arch}
	
	\subsubsection{Dense Spatial Modulation with LUT Transformation} 
	
	Specifically, given an incorrectly exposed input image $x\in\mathbb{R}^{H\times W\times 3}$, we employ a Spatial Signal Encoder~\cite{restormer} $\mathcal{E}$ to extract context-aware spatial features and predict dense modulation weights $\mathcal{M}$:
	\begin{equation}
		\mathcal{M}=\mathcal{E}(x), \qquad\mathcal{M}_{i,j}=[m_{i,j}^{1},m_{i,j}^{2}, \dots, m_{i,j}^{K}],
	\end{equation}
	where $K$ denotes the number of transformation bases. Each vector $\mathcal{M}_{i,j}$ represents the coefficients at location $(i,j)$ in the modulation field. In this way, instead of relying on a single globally shared modulation factor, the network predicts a spatially varying modulation map that can describe region-dependent correction demands more flexibly.
	
	Guided by these estimated weights, we perform image transformation using multiple 3D look-up tables (LUTs)~\cite{3dlut}. Let $\{LUT_{k}\}_{k=1}^{K}$ denote a bank of $K$ learnable 3D LUTs. For an input pixel $x_{i,j}$, the $k$-th LUT yields a transformed value $LUT_{k}(x_{i,j})$ via trilinear interpolation. The corrected result is computed as:
	\begin{equation}
		\tilde{y}_{i,j} = \sum_{k=1}^{K} m_{i,j}^{k}\,LUT_{k}(x_{i,j}).
	\end{equation}
	
	The necessity of our architectural choice can be understood through the Lipschitz constant~\cite{2018lipschitz}, which measures the maximum rate of change of a mapping, \emph{i.e.}, how strongly the output may respond to a slight input perturbation. Exposure correction is fundamentally governed by this regularity property, since coexisting under-, over-, and properly exposed regions within a single scene demand highly divergent transformation magnitudes, requiring the underlying mapping to sustain sharply varying local sensitivities. For a standard deep neural network $\mathcal{F}_\theta$, the overall Lipschitz constant is upper-bounded by the product of the spectral norms of the weight matrices $\{w^{(d)}\}_{d=1}^{D}$~\cite{l-normalized}:
	\begin{equation}
		L_{\mathcal{F}_\theta} \le \prod_{d=1}^{D} \|w^{(d)}\|_2.
	\end{equation}
	When attempting to fit abrupt, fine-grained spatial variations, a monolithic neural network must amplify its internal weights, incurring a heavy penalty from weight decay during training. To avoid this regularization penalty, the network tends to default to learning smooth, global features with small Lipschitz constants~\cite{Orthogonal-Convolutional}, yielding an overly coarse modulation signal that struggles with pixel-level non-uniformity. This phenomenon also offers an explanation for the globally biased behavior of existing methods.
	
	In contrast, our decoupled design avoids this bottleneck by explicitly delegating complex color mapping to the 3D LUT bases. The encoder merely predicts the linear combination weights $\mathcal{M}$, thereby reducing the overall Lipschitz upper bound to:
	\begin{equation}
		L_{\mathrm{overall}} \le \|\mathcal{M}\| \cdot L_{\mathrm{LUT}},
	\end{equation}
	where $L_{\mathrm{LUT}}$ denotes the Lipschitz constant of the LUT bases. Since each LUT performs trilinear interpolation over a finite lattice on the compact color cube, $L_{\mathrm{LUT}}$ is a bounded constant determined solely by the lattice values~\cite{rudin}. Consequently, the LUT bases absorb the steepness of the color transform, alleviating the pressure on the network weights.
	Moreover, although predicting dense spatial weight maps introduces a modest memory overhead, the inherent lookup efficiency of 3D LUTs ensures fast inference, as evidenced by the inference times reported in Table~\ref{tab:lcdp_reed_compare}.

	\subsubsection{Color Compensation Branch}
	
	In addition to the main modulation branch, we introduce a lightweight compensation branch based on the Hue-Saturation-Lightness (HSL) color space as a refinement stage. Since luminance and chromatic information are strongly coupled in sRGB space, performing exposure correction directly in this coupled space often leads to visible color distortion. Some recent methods transform images into Hue-Saturation-Value (HSV)-like spaces to decouple brightness from color~\cite{hvi,cidnet+,iclr}. However, the saturation component in such representations is compressed in over-exposed regions, which cause latent chromatic-information loss. To mitigate this issue, we adopt HSL as an auxiliary color-compensation domain. Compared to HSV-style representations, HSL preserves richer chromatic information under high-brightness conditions, making it significantly more robust for general exposure recovery.
	
	\begin{figure}[t]
		\centering
		\includegraphics[width=\linewidth]{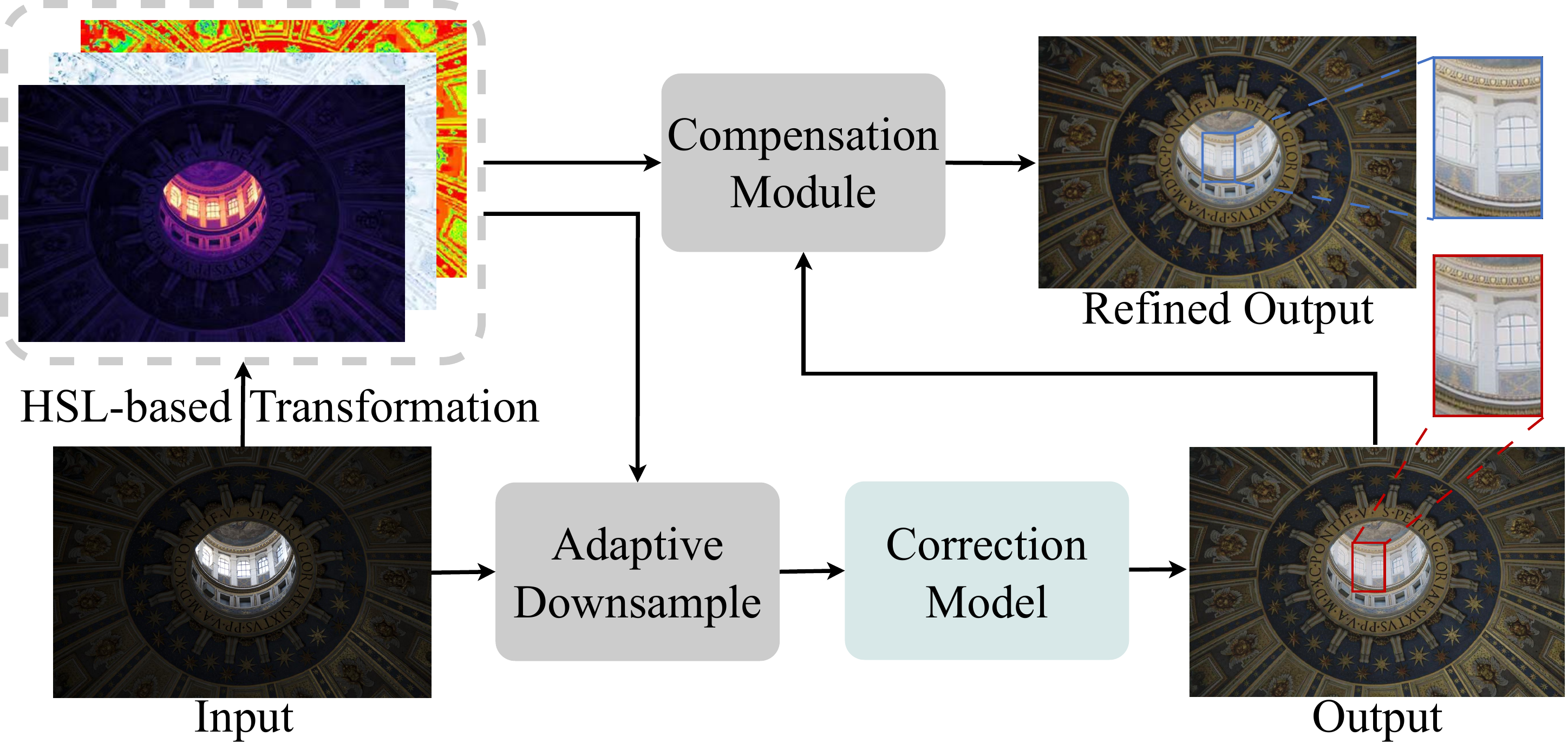}
		\caption{
			Illustration of the proposed HSL-based compensation branch.
			The HSL representation serves two primary purposes. It provides color-aware cues to the adaptive downsampling module and is also fed into a compensation network, where it is fused with the LUT-corrected result to produce the final refined output. 
		}
		\label{fig:hsl_comp}
	\end{figure}
	
	To further alleviate the discontinuity of the hue channel, we polarize the hue representation. Let the hue value be $H\in[0,360^\circ)$, and let $S$ and $L$ denote the saturation and lightness components. We project the hue into a continuous polar coordinate system as
	\begin{equation}
		u=\cos\left(\frac{\pi H}{180}\right),
		\quad
		v=\sin\left(\frac{\pi H}{180}\right).
	\end{equation}
	Based on this, we define a three-channel HSL-derived representation $z$ as
	\begin{equation}
		z
		=
		\left[
		z_{1},z_{2},z_{3}
		\right]
		=
		\left[
		u\cdot S,\; v\cdot S,\; L
		\right],
	\end{equation}
	where the first two channels encode hue and saturation in a continuous polarized form while the third channel preserves the lightness information. 
	
	As illustrated in Fig.~\ref{fig:hsl_comp}, the HSL representation is exploited in two complementary ways. First, inspired by recent LUT-based architectures~\cite{adaint,cotf}, it is integrated into the adaptive downsampling module to provide color-aware cues for predicting the spatial sampling intervals. Second, it is fed into a lightweight compensation network and fused with the LUT-corrected result $\tilde{y}$ to yield the final refined output, \emph{i.e.},
	\begin{equation}
		\hat{y}=\mathcal{C}(z,\tilde{y}),
	\end{equation}
	where $\mathcal{C}(\cdot)$ denotes the compensation module. Within $\mathcal{C}(\cdot)$, the HSL representation $z$ is first encoded into color-correction features. The two branches are then fused via channel-wise cross-attention~\cite{restormer}, which extracts queries from the HSL-based compensation features and derives keys and values from the LUT features.

	\begin{figure*}[t]
		\centering
		\includegraphics[width=\linewidth]{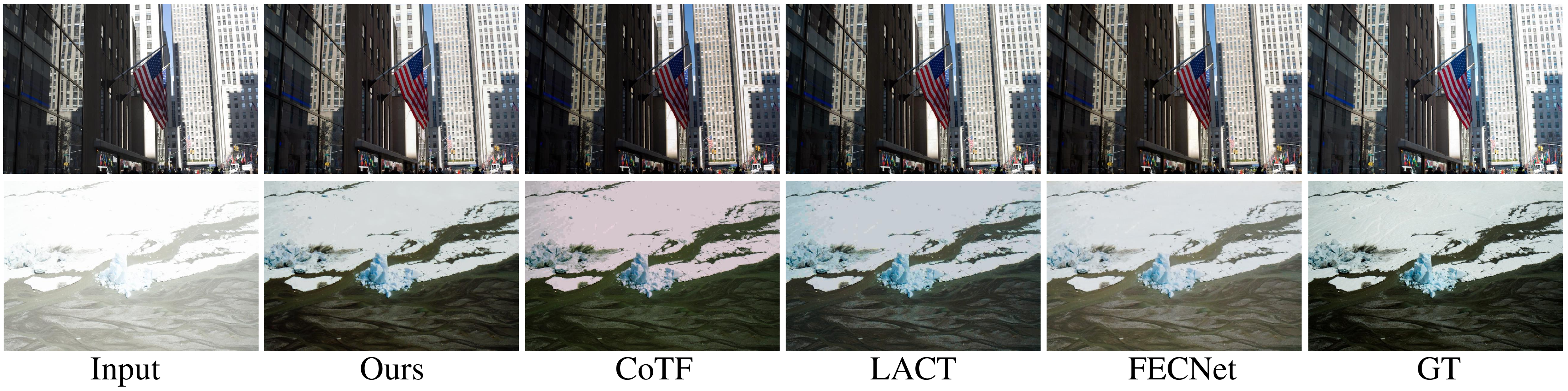}
		\caption{
			Visual comparisons with state-of-the-art methods on the MSEC dataset \cite{msec}. 
		}
		\label{fig:msec_comp}
	\end{figure*}

	\subsection{Uncertainty-inspired Non-uniform Loss}
	\label{sec:l1l2}
	
	Drawing inspiration from uncertainty modeling techniques in low-level vision~\cite{ningqian,fangzhenxuan}, we revisit the loss design for exposure correction from a probabilistic perspective. Let the input be an incorrectly exposed image $x\in\mathbb{R}^{H\times W\times 3}$, and let the exposure correction network $f_\theta(\cdot)$ produce the restored output $\hat{y}=f_\theta(x)$ with supervision target $y$. The residual is defined as
	\begin{equation}
		r = y-\hat{y}.
	\end{equation}
	
	We first show why the standard $L_1$ loss is insufficient for non-uniform exposure correction. Assume that the residuals follow a homoscedastic Laplace distribution:
	\begin{equation}
		r_{i,c}\sim \mathrm{Laplace}(0,b),
		\qquad
		\forall i\in\{1,\dots,HW\},\; c\in\{1,2,3\},
	\end{equation}
	where $b$ is a single global scale parameter shared by all pixels and channels. Since the Laplace density is $p(r)=\frac{1}{2b}\exp\left(-|r|/b\right)$, the negative log-likelihood is
	\begin{equation}
		\mathcal{L}_{\mathrm{NLL}}(\theta;b)
		=
		\sum_{i,c}
		\left(
		\frac{|y_{i,c}-\hat{y}_{i,c}|}{b}
		+ \log b
		\right).
	\end{equation}
	When $b$ is treated as a fixed constant, minimizing $\mathcal{L}_{\mathrm{NLL}}$ is equivalent to minimizing
	$\|y-\hat{y}\|_1$, which recovers the standard $L_1$ loss. For completeness, the Gaussian counterpart is
	\begin{equation}
		r_{i,c}\sim\mathcal{N}(0,\sigma^2),\qquad
		\mathcal{L}_{\mathrm{NLL}}^{G}=\sum_{i,c}
		\left(\frac{r_{i,c}^2}{2\sigma^2}+\frac{1}{2}\log\sigma^2\right).
	\end{equation}
	With fixed $\sigma$, minimizing this objective is equivalent to the standard $L_2$ loss. Thus, conventional $L_1$ and $L_2$ reconstruction losses are both implicitly homoscedastic: all pixels are optimized under a shared global uncertainty scale.
	
	To properly model the spatially varying correction demands, the assumption of a single global scale must be relaxed. Instead, we assign each pixel its own uncertainty: 
	\begin{equation}
		r_{i,c}\sim \mathrm{Laplace}(0,b_{i,c}),
		\qquad
		b_{i,c}>0,
	\end{equation}
	where $b_{i,c}$ denotes the pixel-wise uncertainty scale. 
	
	Following Ning \emph{et al.}~\cite{ningqian}, we model the joint distribution of $r_{i,c}$ and $b_{i,c}$ conditioned on $x_{i,c}$ to constrain the network to generate meaningful uncertainty estimates. The conditional probability is factorized as:
	\begin{equation} 
		p(r_{i,c}, b_{i,c} \mid x_{i,c}) = p(r_{i,c} \mid b_{i,c} , x_{i,c}) p(b_{i,c}). 
	\end{equation}
	While previous uncertainty formulations often rely on sparsity-promoting priors~\cite{jeffreysPrior}, exposure errors typically manifest as dense, spatially continuous degradations. Accordingly, we adopt an uninformative prior $p(b_{i,c}) \propto 1$, which does not impose a sparsity bias and therefore permits dense, region-wise uncertainty estimates.

	For numerical stability, we parameterize the uncertainty in the log-space as $s_{i,c}=\log b_{i,c}$.  The final uncertainty-inspired non-uniform (UNU) loss is thus derived by minimizing the conditional negative log-likelihood:
	\begin{equation}
		\mathcal{L}_{UNU}
		=
		\sum_{i,c}
		\left(
		e^{-s_{i,c}}|y_{i,c}-\hat{y}_{i,c}| + s_{i,c}
		\right).
	\end{equation}
	
	In practice, the log-scale field $s$ is predicted by a lightweight Uncertainty Estimator, which takes the model output as input and whose architecture is illustrated in Fig.~\ref{fig:framework}. The estimator is trained jointly with the correction network through $\mathcal{L}_{UNU}$.
	
\begin{figure}[t]
		\centering
		\includegraphics[width=\linewidth]{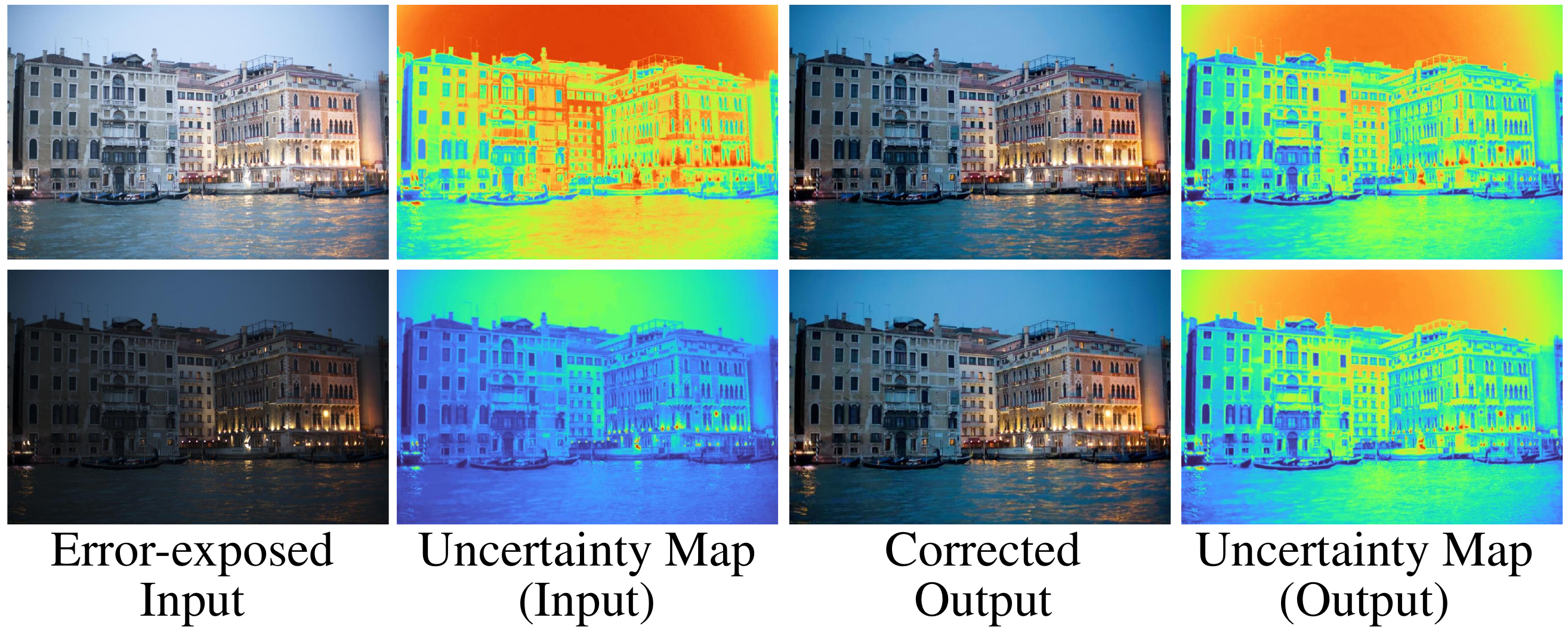}
		\caption{
			Visualization of the predicted uncertainty maps using different inputs to the Uncertainty Estimator. As shown, the estimated uncertainty successfully distinguishes between the two distinct degradation patterns, validating its role in dynamically balancing regional optimization weight.
		}
		\label{fig:uncertainty_abla}
	\end{figure}

	This formulation enables the network to adaptively allocate supervision strength according to regional correction demands, providing a well-suited optimization objective for spatially non-uniform exposure correction. Notably, the predicted residual scale in our formulation is not interpreted as a conventional indicator of restoration difficulty. Because real-world exposure correction involves two opposite degradation directions, directly equating a larger scale with higher recovery difficulty is inappropriate. Instead, this scale functions as a dense spatial reweighting signal that dynamically balances supervision across heterogeneous regions. As visualized in Fig.~\ref{fig:uncertainty_abla}, the uncertainty maps predicted for the same scene under under-exposure ($-1.5$ EV) and over-exposure ($+1.5$ EV) exhibit distinct responses, with high uncertainty (warm colors) concentrated in over-exposed regions and low uncertainty (cold colors) in under-exposed regions. After correction, the uncertainty across these regions converges toward intermediate values, indicating that the estimator adaptively tracks where the strongest corrections are needed. The effectiveness of this mechanism is further analyzed in Sec.~\ref{abla_uncertainty}.
	
	Alongside our $\mathcal{L}_{UNU}$, we employ the standard $L_1$ loss $\mathcal{L}_{1}$, perceptual loss $\mathcal{L}_{per}$, and SSIM loss $\mathcal{L}_{ssim}$ to jointly train the network. The overall objective function is expressed as:
	\begin{equation}
		\mathcal{L}_{total} = \mathcal{L}_{1} + \beta_1 \mathcal{L}_{per} + \beta_2 \mathcal{L}_{ssim} + \beta_3 \mathcal{L}_{UNU},
	\end{equation}
	where $\beta_1$, $\beta_2$, and $\beta_3$ are empirically set to $0.1$, $0.1$, and $0.1$.
	
	\begin{table*}[t]
		\caption{Quantitative comparisons on MSEC \cite{msec} and SICEV2 \cite{enc}. The best and second-best results are highlighted in \textcolor{red}{\textbf{red}} and \textcolor{blue}{\textbf{blue}}, respectively. All results are obtained either from the officially released model weights of the compared methods or as reported in their original papers. ``--'' denotes unavailable results.}
		\label{tab:msec_sice_compare}
		\centering
		\resizebox{\textwidth}{!}{
			\begin{tabular}{lccccccc ccccccc}
				\toprule[1.5pt]
				\multirow{3}{*}{\textbf{Methods}}
				& \multicolumn{7}{c}{\textbf{MSEC}}
				& \multicolumn{7}{c}{\textbf{SICEV2}} \\
				\cmidrule(lr){2-8} \cmidrule(lr){9-15}
				& \multicolumn{2}{c}{\textbf{Under}}
				& \multicolumn{2}{c}{\textbf{Over}}
				& \multicolumn{3}{c}{\textbf{Average}}
				& \multicolumn{2}{c}{\textbf{Under}}
				& \multicolumn{2}{c}{\textbf{Over}}
				& \multicolumn{3}{c}{\textbf{Average}} \\
				\cmidrule(lr){2-3} \cmidrule(lr){4-5} \cmidrule(lr){6-8}
				\cmidrule(lr){9-10} \cmidrule(lr){11-12} \cmidrule(lr){13-15}
				& PSNR$\uparrow$ & SSIM$\uparrow$
				& PSNR$\uparrow$ & SSIM$\uparrow$
				& PSNR$\uparrow$ & SSIM$\uparrow$ & LPIPS$\downarrow$
				& PSNR$\uparrow$ & SSIM$\uparrow$
				& PSNR$\uparrow$ & SSIM$\uparrow$
				& PSNR$\uparrow$ & SSIM$\uparrow$ & LPIPS$\downarrow$ \\
				\midrule
				ZeroDCE (CVPR20)    
				& 14.55 & 0.5887 & 10.40 & 0.5142 & 12.06 & 0.5441 & 0.2923 & 16.92 & 0.6330 &  7.11 & 0.4292 & 12.02 & 0.5311 & 0.3532 \\
				RUAS (CVPR21)        
				& 13.43 & 0.6807 &  6.39 & 0.4655 &  9.20 & 0.5515 & 0.4819 & 16.63 & 0.5589 &  4.54 & 0.3196 & 10.59 & 0.4393 & 0.5122 \\
				SCI (CVPR22)        
				&  9.97 & 0.6681 &  5.83 & 0.5190 &  7.49 & 0.5786 & 0.3116 & 17.86 & 0.6401 &  4.45 & 0.3629 & 12.49 & 0.5051 & 0.4239 \\
				PairLIE (CVPR23)      
				& 11.78 & 0.6596 &  8.37 & 0.5887 &  9.73 & 0.6171 & 0.3605 & 16.67 & 0.5995 &  6.26 & 0.3846 & 11.47 & 0.4921 & 0.4138 \\
				MSEC (CVPR21)        
				& 20.52 & 0.8129 & 19.79 & 0.8156 & 20.08 & 0.8145 & 0.1721 & 19.62 & 0.6512 & 17.59 & 0.6560 & 18.58 & 0.6536 & 0.2814 \\
				FECNet (ECCV22)      
				& 22.96 & 0.8598 & 23.22 & 0.8748 & 23.12 & 0.8688 & 0.1419 & 22.01 & 0.6737 & 19.91 & 0.6961 & 20.96 & 0.6849 & 0.2656 \\
				LCDPNet (ECCV22)  
				& 22.35 & \textcolor{blue}{\textbf{0.8650}} & 22.17 & 0.8476 & 22.30 & 0.8552 & 0.1451 & 17.45 & 0.5622 & 17.04 & 0.6463 & 17.25 & 0.6043 & 0.2592 \\
				ERL (CVPR23)  
				& 23.10 & 0.8639 & 23.18 & 0.8759 & 23.15 & 0.8711 & --     & 22.35 & 0.6671 & 20.10 & 0.6891 & 21.22 & 0.6781 & --     \\
				MMHT (MM23)
				& 22.97 & 0.8560 & 23.10 & 0.8709 & 23.05 & 0.8650
				& -- & 22.55 & 0.7090 & \textcolor{blue}{\textbf{21.06}} & 0.7237 & \textcolor{blue}{\textbf{21.81}} & 0.7164 & -- \\
				LACT (ICCV23) 
				& \textcolor{blue}{\textbf{23.49}} & 0.8620 & \textcolor{blue}{\textbf{23.68}} & 0.8720 & \textcolor{blue}{\textbf{23.57}} & 0.8690
				& 0.1239 & 22.35 & \textcolor{blue}{\textbf{0.7102}} & 20.54 & 0.7197 & 21.45 & 0.7150 & 0.2678 \\
				
				UEC (ECCV24)
				& 18.79 & 0.7945 & 19.34 & 0.8119 & 19.07 & 0.8044
				& 0.1852 & 17.26 & 0.6720 & 16.95 & 0.6896 & 17.10 & 0.6808 & 0.2198 \\
				CSEC (CVPR24)
				& 22.10 & 0.8453 & 22.08 & 0.8601 & 22.16 & 0.8549
				& 0.1495 & 19.68 & 0.6736 & 19.46 & 0.7203 & 19.57 & 0.6970 & 0.2260 \\
				CoTF (CVPR24) 
				& 23.36 & 0.8630 & 23.49 & \textcolor{blue}{\textbf{0.8793}} & 23.44 & \textcolor{blue}{\textbf{0.8728}} & \textcolor{blue}{\textbf{0.1232}} & \textcolor{blue}{\textbf{22.90}} & 0.7029 & 20.13 & \textcolor{blue}{\textbf{0.7274}} & 21.51 & 0.7151 & \textcolor{blue}{\textbf{0.1924}} \\
				
				CLIER (ICCV25)
				& -- & -- & -- & -- & 22.03 & 0.8481
				& -- & -- & -- & -- & -- & 20.19 & \textcolor{red}{\textbf{0.8141}} & -- \\
				LoopExpose (TIP26)
				& 21.15 & 0.8276 & 20.92 & 0.8375 & 21.06 & 0.8347
				& 0.1630 & 16.94 & 0.6239 & 16.25 & 0.6524 & 16.59 & 0.6382 & 0.2502 \\
				\textbf{Ours}
				& \textcolor{red}{\textbf{23.64}} & \textcolor{red}{\textbf{0.8721}} & \textcolor{red}{\textbf{23.83}} & \textcolor{red}{\textbf{0.8853}} & \textcolor{red}{\textbf{23.72}} & \textcolor{red}{\textbf{0.8774}} & \textcolor{red}{\textbf{0.1197}} 
				& \textcolor{red}{\textbf{23.49}} & \textcolor{red}{\textbf{0.7146}} & \textcolor{red}{\textbf{21.85}} & \textcolor{red}{\textbf{0.7422}} & \textcolor{red}{\textbf{22.67}} & \textcolor{blue}{\textbf{0.7284}} & \textcolor{red}{\textbf{0.1847}} \\
				\bottomrule[1.5pt]
			\end{tabular}
		}
	\end{table*}
	
	\begin{figure*}[t]
		\centering
		\includegraphics[width=\linewidth]{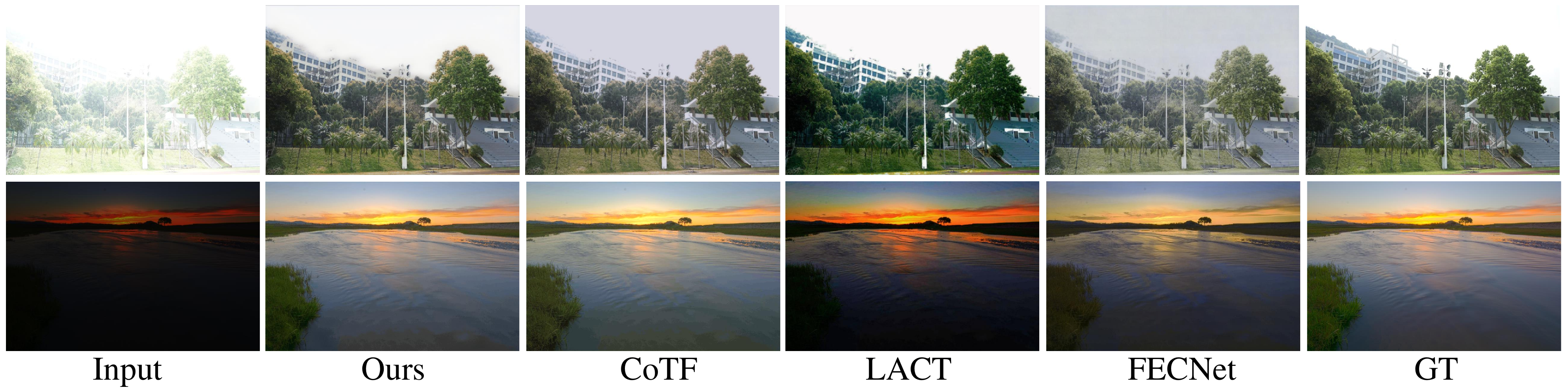}
		\caption{
			Visual comparisons with state-of-the-art methods on the SICEV2 dataset \cite{enc}. 
		}
		\label{fig:sice_comp}
	\end{figure*}

	\section{Experiments}

	\subsection{Settings}
	
	\subsubsection{Datasets}
	
	We evaluate our proposed method against state-of-the-art baselines on four exposure correction benchmarks: three widely adopted datasets (MSEC~\cite{msec}, SICEV2~\cite{enc}, and LCDP~\cite{lcdp}), alongside a recently proposed real-world benchmark (REED~\cite{reed}). The MSEC dataset is constructed by rendering raw images from MIT-Adobe FiveK~\cite{fivek} with relative exposure values (EVs) ranging from $-1.5$ to $+1.5$. For the real-captured SICE dataset \cite{sice}, which features sequences with fixed EV steps, we follow the evaluation protocol established in~\cite{enc}. The LCDP dataset~\cite{lcdp} is generated following a pipeline similar to MSEC, but is specifically tailored for non-uniform exposure correction. Finally, the REED dataset~\cite{reed} is collected using a camera's burst-shooting mode to capture real-world scenes with relative EVs ranging from $-3$ to $+3$.

	\subsubsection{Implementation Details}
	
	During training, input images are cropped into patches of size $768 \times 768$ with a batch size of $1$. We optimize the network using Adam~\cite{adam} ($\beta_1 = 0.9$, $\beta_2 = 0.99$), where the learning rate is scheduled via cosine annealing from $4 \times 10^{-4}$ down to $1 \times 10^{-6}$. 
	All experiments are conducted on a single NVIDIA RTX 4090 GPU. Additional implementation details are available in our code repository.
	
	\begin{table*}[t]
		\centering
		
		\caption{Quantitative comparisons on LCDP~\cite{lcdp} and REED~\cite{reed}. The best and second-best results are highlighted in \textcolor{red}{\textbf{red}} and \textcolor{blue}{\textbf{blue}}, respectively. All results are obtained either from the officially released model weights of the compared methods or as reported in the original papers.  ``--'' denotes unavailable results.}
		\label{tab:lcdp_reed_compare}
		\resizebox{\textwidth}{!}{
			\begin{tabular}{lcccccccccccc}
				\toprule[1.5pt]
				\multirow{3}{*}{\textbf{Methods}}
				& \multicolumn{3}{c}{\textbf{LCDP}}
				& \multicolumn{8}{c}{\textbf{REED}}
				& \multirow{3}{*}{\textbf{\makecell[c]{Inference \\ Time (s)}}$\downarrow$} \\
				\cmidrule(lr){2-4}
				\cmidrule(lr){5-12}
				& \multicolumn{3}{c}{\textbf{Average}}
				& \multicolumn{2}{c}{\textbf{Under}}
				& \multicolumn{2}{c}{\textbf{Over}}
				& \multicolumn{4}{c}{\textbf{Average}}
				& \\
				\cmidrule(lr){2-4}
				\cmidrule(lr){5-6}
				\cmidrule(lr){7-8}
				\cmidrule(lr){9-12}
				& \textbf{PSNR}$\uparrow$ & \textbf{SSIM}$\uparrow$ & \textbf{LPIPS}$\downarrow$
				& \textbf{PSNR}$\uparrow$ & \textbf{SSIM}$\uparrow$
				& \textbf{PSNR}$\uparrow$ & \textbf{SSIM}$\uparrow$
				& \textbf{PSNR}$\uparrow$ & \textbf{SSIM}$\uparrow$ & \textbf{LPIPS}$\downarrow$ & \textbf{NIQE}$\downarrow$
				& \\
				\midrule
				ZeroDCE++ (TPAMI21) & 18.42 & 0.7669 & 0.2204 & 12.58 & 0.6503 & 13.02 & 0.6847 & 12.80 & 0.6675 & 0.19 & 4.55 & \textcolor{blue}{\textbf{0.0024}} \\
				FECNet (ECCV22) & 22.34 & 0.8038 & 0.2334 & 16.68 & 0.6823 & 13.93 & 0.5881 & 15.31 & 0.6352 & 0.33 & 3.99 & 0.1261 \\
				LCDP (ECCV22) & 23.24 & 0.8420 & 0.1368 & 16.75 & 0.7779 & 17.01 & 0.8139 & 16.88 & 0.7959 & 0.12 & 5.00 & 0.0472 \\
				SCI (CVPR22) & 15.96 & 0.6646 & 0.2913 & 16.99 & 0.7436 & 15.25 & 0.7034 & 16.12 & 0.7235 & 0.14 & 5.07 & \textcolor{red}{\textbf{0.0021}} \\
				CoTF (CVPR24) &  \textcolor{blue}{\textbf{23.89}} & \textcolor{blue}{\textbf{0.8581}} & \textcolor{blue}{\textbf{0.1035}} & 17.84 & 0.7468 & 18.22 & 0.7546 & 18.03 & 0.7518 & 0.18 & 5.39 & 0.0095 \\
				UEC (ECCV24) & 15.79 & 0.6352 & 0.3307 & 17.47 & 0.8141 & 17.57 & 0.7768 & 17.52 & 0.7954 & 0.18 & \textcolor{blue}{\textbf{3.47}} & 0.0889 \\
				CLIER (ICCV25) & -- & -- & -- & 19.63 & \textcolor{blue}{\textbf{0.8546}} & \textcolor{blue}{\textbf{18.97}} & \textcolor{blue}{\textbf{0.8470}} & \textcolor{blue}{\textbf{19.25}} & \textcolor{blue}{\textbf{0.8503}} & \textcolor{red}{\textbf{0.09}} & 3.74 & -- \\
				LoopExpose (TIP26) & 17.22 & 0.6366 & 0.2855 & \textcolor{blue}{\textbf{19.93}} & 0.8539 & 18.53 & 0.8125 & 19.23 & 0.8332 & 0.16 & 3.89 & 0.0113 \\
				\midrule
				\textbf{Ours} & \textcolor{red}{\textbf{24.41}} & \textcolor{red}{\textbf{0.8679}} & \textcolor{red}{\textbf{0.0950}} & \textcolor{red}{\textbf{26.91}} & \textcolor{red}{\textbf{0.9324}} & \textcolor{red}{\textbf{24.53}} & \textcolor{red}{\textbf{0.9157}} & \textcolor{red}{\textbf{25.68}} & \textcolor{red}{\textbf{0.9241}} & \textcolor{blue}{\textbf{0.10}} & \textcolor{red}{\textbf{3.31}} & 0.0566 \\
				\bottomrule[1.5pt]
		\end{tabular}}
	\end{table*}

	\subsection{Comparison with State-of-the-art Methods}
	
	Methods participating in our comparison include MSEC~\cite{msec}, FECNet~\cite{fecnet}, LCDPNet~\cite{lcdp}, ERL \cite{erl}, MMHT \cite{mmht}, LACT~\cite{lact}, UEC~\cite{uec}, CSEC \cite{csec}, CoTF~\cite{cotf}, CLIER~\cite{reed}, LoopExpose~\cite{loopexpose}, ZeroDCE~\cite{zerodce}, ZeroDCE++\cite{zerodce++}, RUAS\cite{ruas}, SCI~\cite{sci}, and PairLIE~\cite{pairlie}.

	\subsubsection{Quantitative Comparison}
	
	To ensure a fair and comprehensive evaluation, we adopt PSNR \cite{psnr}, SSIM~\cite{ssim}, LPIPS~\cite{lpips}, and NIQE~\cite{niqe} as our quantitative metrics. As summarized in Tables~\ref{tab:msec_sice_compare} and \ref{tab:lcdp_reed_compare}, our method consistently achieves state-of-the-art or highly competitive performance across the MSEC \cite{msec}, SICEV2 \cite{enc}, LCDP \cite{lcdp}, and REED \cite{reed} benchmarks. These results clearly demonstrate the effectiveness of explicitly modeling spatially non-uniform correction demands. Notably, our model yields the most significant performance gains on the LCDP dataset, which is specifically tailored for non-uniform scenarios. Furthermore, our approach maintains favorable computational efficiency compared to recent high-performing baselines, striking a compelling balance between restoration fidelity and computational cost.

	\begin{figure*}[t]
		\centering
		\includegraphics[width=\linewidth]{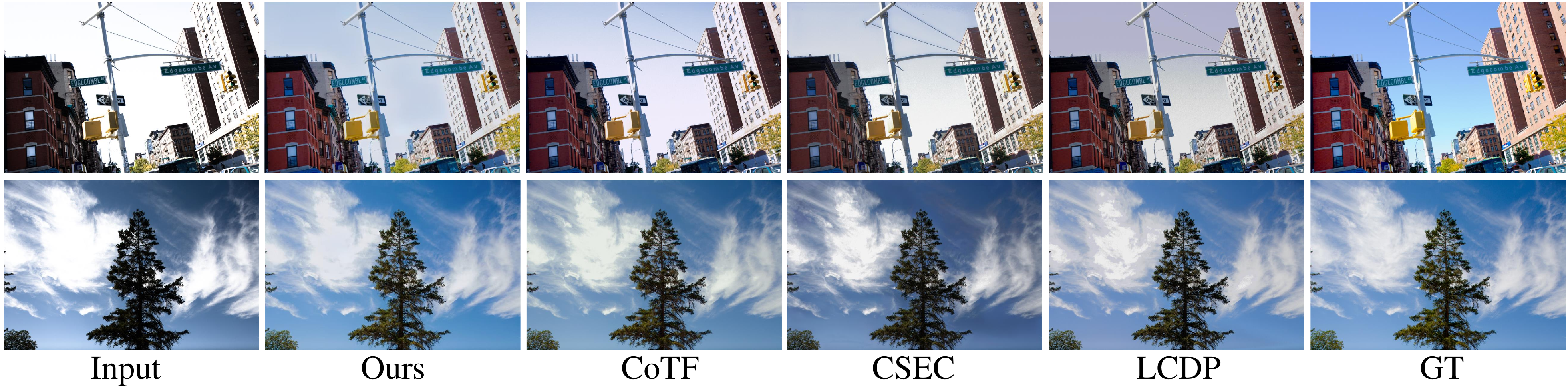}
		\caption{
			Visual comparisons with state-of-the-art methods on the LCDP dataset \cite{lcdp}. 
		}
		\label{fig:lcdp_comp}
	\end{figure*}
	
	\subsubsection{Qualitative Comparison}
	
	Visual comparisons are shown in Figs.~\ref{fig:msec_comp}, \ref{fig:sice_comp}, and~\ref{fig:lcdp_comp}. While existing methods often exhibit globally biased correction behaviors, our spatially adaptive framework restores the full dynamic range more faithfully with high color fidelity. These qualitative results agree with the quantitative findings and support our analysis of the limitations of uniform correction assumptions in complex scenes.
	
	\subsection{Unsupervised Training}
	\label{sec:unsupervised_training}
	
	We further evaluate our method under unsupervised training, following the setting of LoopExpose~\cite{loopexpose}, using the fused high dynamic range image for pseudo-target. Under this protocol, our model achieves a mean PSNR of $28.03$ dB against the fusion pseudo-targets, surpassing the officially released checkpoint of LoopExpose ($25.17$ dB) by $2.85$ dB.
	
\begin{figure}[!t]
		\centering
		\includegraphics[width=\columnwidth]{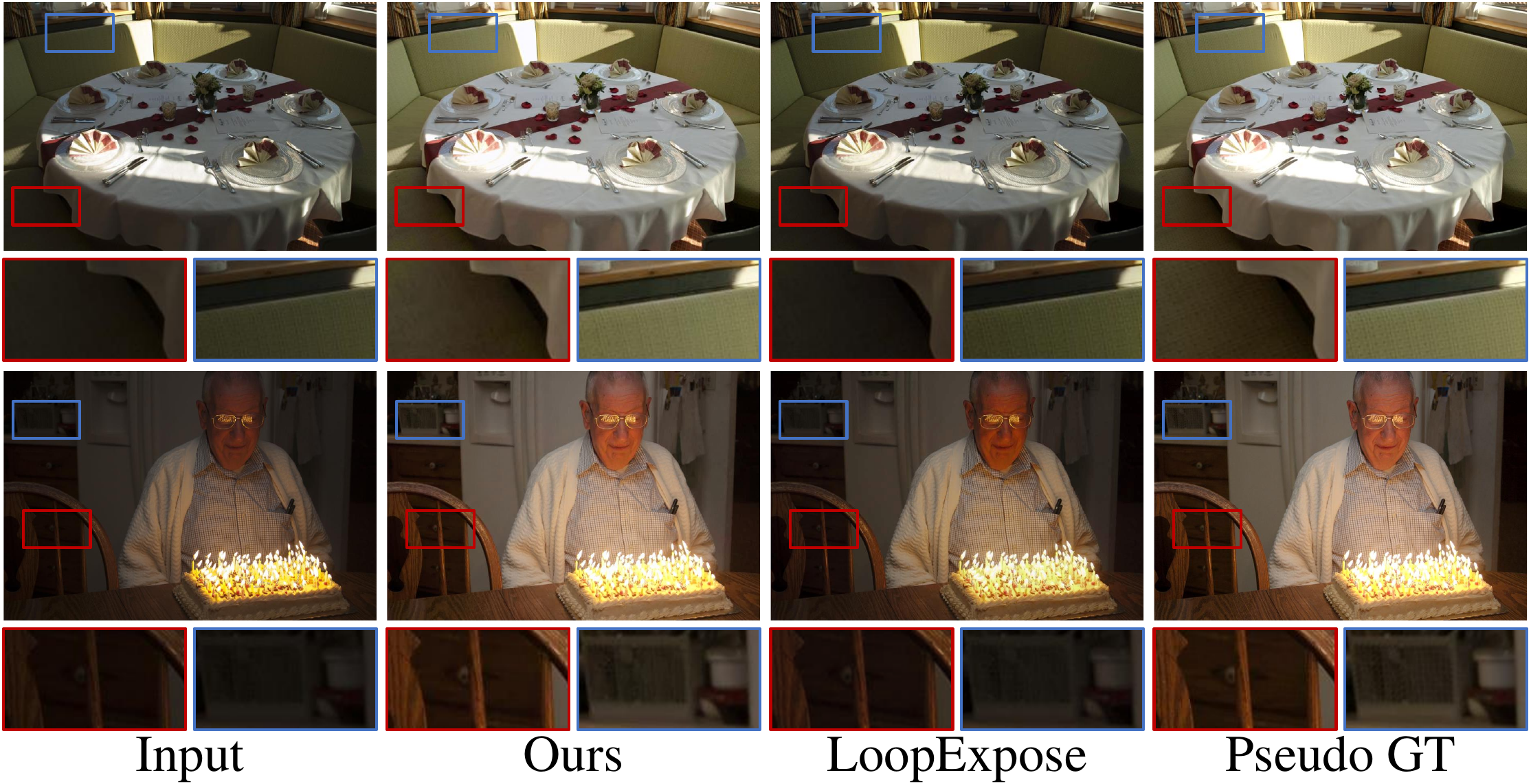}
		\caption{Visual comparison under unsupervised training on MSEC~\cite{msec}. From left to right: input, ours, LoopExpose~\cite{loopexpose}, and the Mertens fusion pseudo-target~\cite{mertens}. Our method faithfully aligns with the fusion pseudo-target by recovering fine shadow details while preserving highlight regions, whereas LoopExpose still suffers from noticeable global bias.}
		\label{fig:unsupervised_training}
	\end{figure}

	This setting complements the supervised benchmarks. Notably, certain ground-truth targets in existing benchmarks still exhibit residual exposure non-uniformity, which may limit both supervised training and full-reference evaluation. In contrast, the fusion-based pseudo-target assembles well-exposed regions from multiple exposures of a scene and covers a broader dynamic range, providing spatially uniform supervision for every region. Recovering such a target from a single degraded input places even greater demands on spatially varying restoration, making the fusion-based protocol a particularly rigorous test of non-uniform correction capability. As illustrated in Fig.~\ref{fig:unsupervised_training}, our model effectively enhances shadow details while preserving highlight regions, yielding regional corrections that closely match the fused pseudo-targets. Together with the gains achieved under expert supervision, this consistent advantage demonstrates that our spatially adaptive framework is not tied to a specific supervision paradigm and can effectively learn from different targets.
	
	\section{Ablation Studies}
	
	We conduct all ablation studies under the identical experimental setup and report the results on the MSEC dataset.
	
	\subsection{Effect of the Color Compensation Design}

	To validate our color compensation design, we investigate the role of the HSL color space in both the adaptive sampling stage and the compensation branch (see Table~\ref{tab:color_comp_ablation}). First, even without the compensation branch (Settings A--C), our baseline achieves strong performance that surpasses nearly all existing methods, demonstrating the standalone effectiveness of our core spatially adaptive framework. Second, across all configurations, adopting the HSL representation consistently outperforms its RGB counterpart, confirming the advantage of HSL-based color modeling for exposure correction.
	
	\begin{table}[t]
		\centering
		\caption{Ablation study of the color compensation design. We analyze the color spaces used in both the adaptive sampling and compensation modules. Here, bilinear denotes the baseline setting that employs standard bilinear downsampling.}
		\label{tab:color_comp_ablation}
		\begin{tabular}{c cc cc cc}
			\toprule[1.5pt]
			\multirow{2}{*}{Setting} 
			& \multicolumn{2}{c}{Sampling} 
			& \multicolumn{2}{c}{Compensation} 
			& \multirow{2}{*}{PSNR$\uparrow$} 
			& \multirow{2}{*}{SSIM$\uparrow$} \\
			\cmidrule(lr){2-3}
			\cmidrule(lr){4-5}
			& RGB & HSL & RGB & HSL & & \\
			\midrule
			A & \multicolumn{2}{c}{bilinear}   &  &  & 23.49  & 0.8732 \\
			B & \checkmark &  &  &  & 23.60 & 0.8744 \\
			C &  & \checkmark &  &  & 23.64 & 0.8751 \\
			D & \checkmark &  & \checkmark &  & 23.62 & 0.8749 \\
			E &  & \checkmark & \checkmark &  & 23.68 & 0.8770 \\
			F & \checkmark &  &  & \checkmark & 23.66 &  0.8762\\ 
			G &  & \checkmark &  & \checkmark & \textbf{23.72} & \textbf{0.8774} \\
			\bottomrule[1.5pt]
		\end{tabular}
	\end{table}

	\subsection{Effect of Different Modulation Strategies}
	\label{sec:modulation_abla}
	
\begin{figure}[t]
		\centering
		\includegraphics[width=\linewidth]{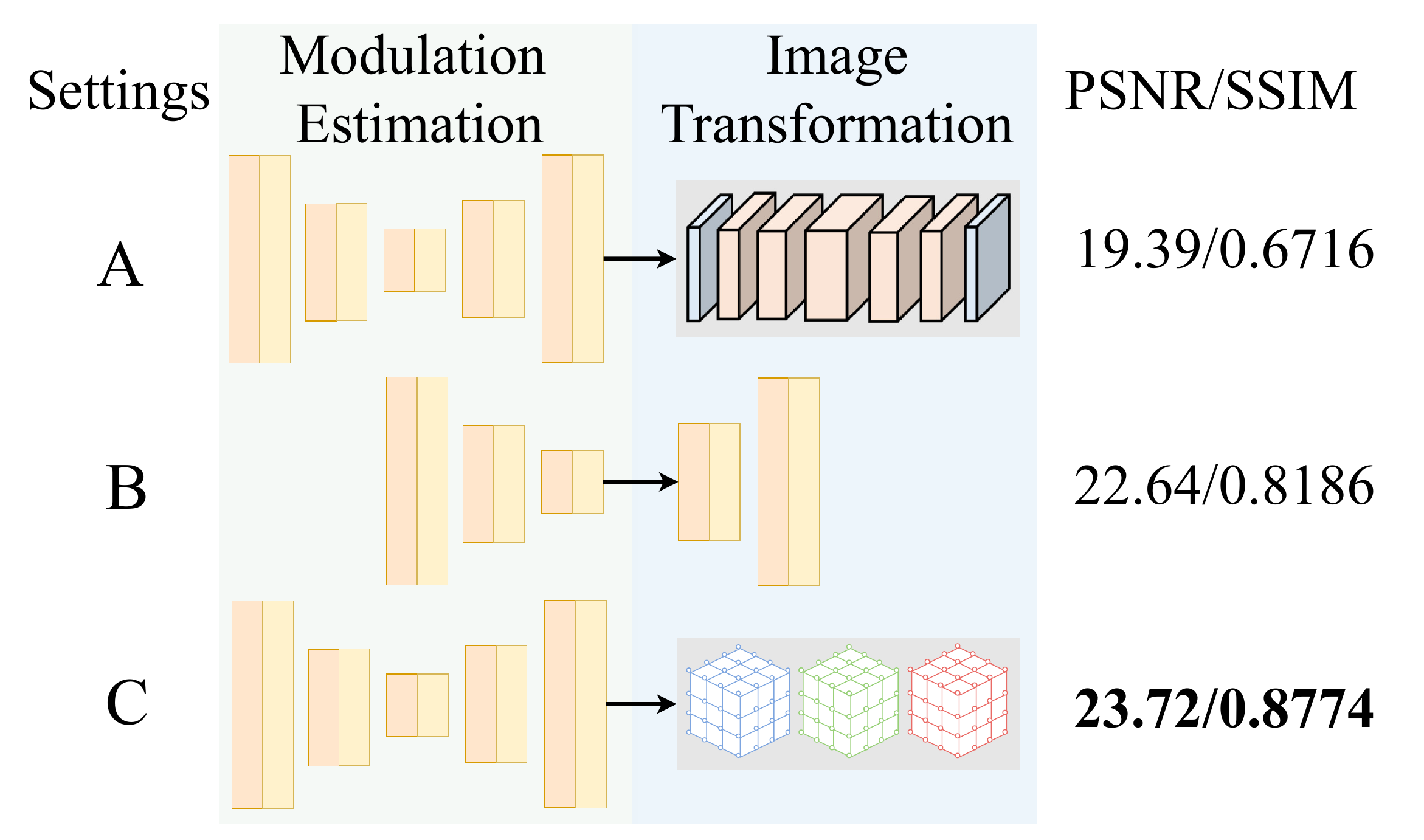}
		\caption{Quantitative and structural comparison of different modulation strategies. Setting A uses plain convolution, Setting B adopts an implicit decoder, and Setting C employs explicit 3D LUT-based transformation. }
		\label{fig:modulation_quant}
	\end{figure}

	To validate our motivation of explicitly decoupling non-uniform modulation estimation from image transformation, we investigate the effect of different modulation strategies. As summarized in Fig.~\ref{fig:modulation_quant}, we construct three variants that share the same backbone but employ different transformation mechanisms: Setting A adopts a plain convolutional transformation following~\cite{lact}, Setting B utilizes an implicit decoder-based transformation, and Setting C employs our explicit 3D LUT-based transformation.

\begin{figure}[t]
		\centering
		\includegraphics[width=\linewidth]{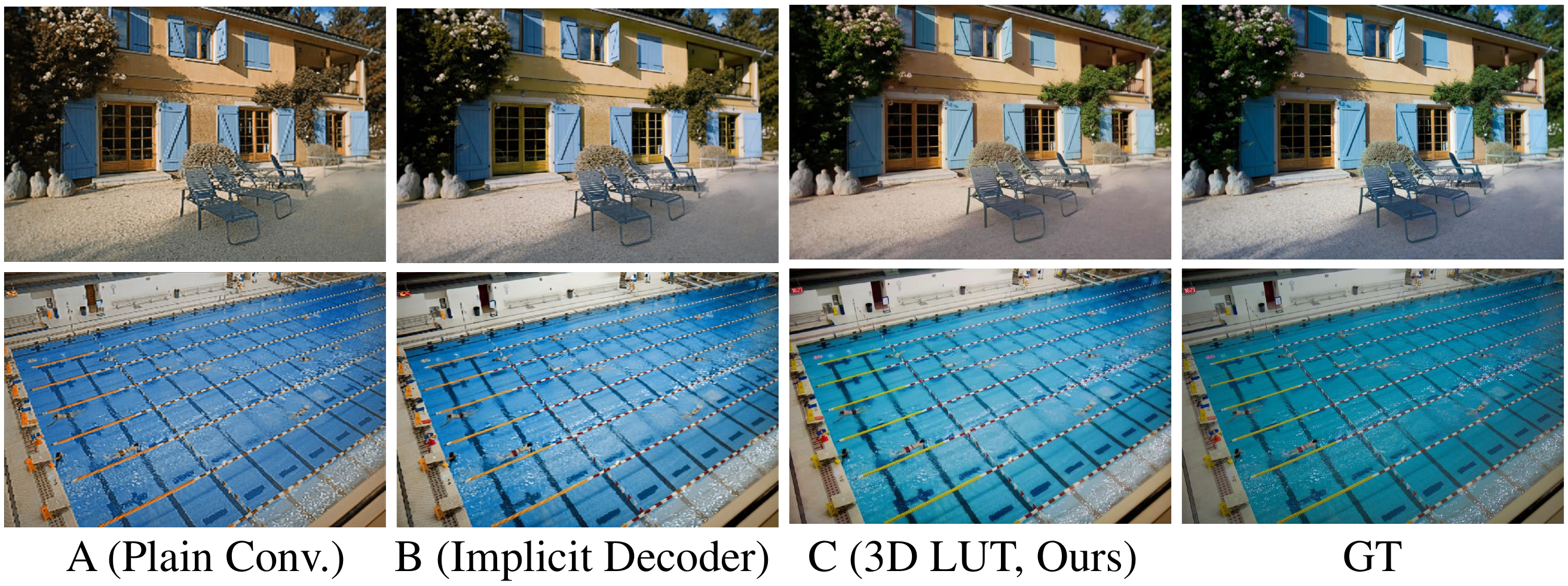}
		\caption{Visual comparison of different modulation strategies. Settings A and B both suffer from biased color distortions, whereas Setting C (Ours) yields the most visually pleasing and accurate restoration.}
		\label{fig:modulation_visual}
	\end{figure}

	The quantitative results in Fig.~\ref{fig:modulation_quant} reveal a substantial performance gap among these strategies. The plain convolutional variant achieves 19.39/0.6716 in PSNR/SSIM, supporting the analysis in Sec.~\ref{sec:arch} that a monolithic mapping is difficult to optimize for fine-grained correction. The implicit decoder (Setting B) extracts an aggregated, near-global modulation signal and improves performance to 22.64 dB and 0.8186. It nevertheless remains below the explicit LUT-based design, indicating that dense, region-adaptive modulation is beneficial under the same backbone. The visual comparisons in Fig.~\ref{fig:modulation_visual} show the corresponding global color bias in Settings A and B.
	
	\subsection{Effect of the Uncertainty-inspired Non-uniform Loss}
	\label{abla_uncertainty}
	
	We further evaluate the proposed uncertainty-inspired non-uniform loss $\mathcal{L}_{UNU}$ in Table~\ref{tab:uncertainty_ablation}. Beyond our own framework, we plug $\mathcal{L}_{UNU}$ into two representative exposure correction architectures, CoTF and FECNet~\cite{fecnet}, by attaching the same lightweight Uncertainty Estimator to their model outputs. The loss consistently improves all three architectures by a comparable margin, indicating that the benefit stems from the spatially reweighted supervision itself rather than from a particular transformation design. Moreover, our framework without $\mathcal{L}_{UNU}$ already surpasses both baselines, and equipping it with $\mathcal{L}_{UNU}$ yields the best overall performance, confirming that the architectural design and the loss design are complementary.

	\begin{table}[t]
		\centering
		\caption{Ablation study of the proposed uncertainty-inspired non-uniform loss ($\mathcal{L}_{UNU}$) on different exposure correction methods. The Uncertainty Estimator takes the model output as input. CoTF~\cite{cotf} and FECNet~\cite{fecnet} are both retrained following their released training setups for a fair comparison.}
		\label{tab:uncertainty_ablation}
		\begin{tabular}{clcc}
			\toprule[1.5pt]
			Setting & Method & PSNR$\uparrow$ & SSIM$\uparrow$ \\
			\midrule
			A & CoTF                       & 23.44 & 0.8728 \\
			B & CoTF w/ $\mathcal{L}_{UNU}$  & 23.56 & 0.8739 \\
			\cmidrule{2-4}
			C & FECNet                     & 22.68 & 0.8557 \\
			D & FECNet w/ $\mathcal{L}_{UNU}$ & 22.80 & 0.8571 \\
			\cmidrule{2-4}
			E & Ours w/o $\mathcal{L}_{UNU}$ & 23.59 & 0.8762 \\
			F & Ours                       & \textbf{23.72} & \textbf{0.8774} \\
			\bottomrule[1.5pt]
		\end{tabular}
	\end{table}

	\section{Conclusion}

This work revisited real-world exposure correction through the lens of spatial non-uniformity, demonstrating that the globally biased behavior of existing methods stems fundamentally from the uniformity assumptions embedded in both network architectures and optimization objectives. By decoupling dense spatial modulation from color transformations, our framework effectively alleviates the Lipschitz bottleneck of monolithic networks. In parallel, relaxing the homoscedastic assumption of standard reconstruction losses through dense uncertainty modeling enables the optimization process to dynamically balance opposing exposure degradations. Comprehensive evaluations across supervised and unsupervised paradigms, together with plug-and-play ablation studies, confirm that jointly modeling spatial non-uniformity in representation and supervision yields substantial improvements over state-of-the-art approaches. In future work, we plan to extend this spatially adaptive paradigm to video exposure correction by incorporating temporal consistency constraints, and explore generative priors to recover textures in severely clipped regions.
	
	\bibliographystyle{IEEEtran}
	\bibliography{main}

	\begin{IEEEbiography}[{\includegraphics[width=1in,height=1.25in,clip,keepaspectratio]{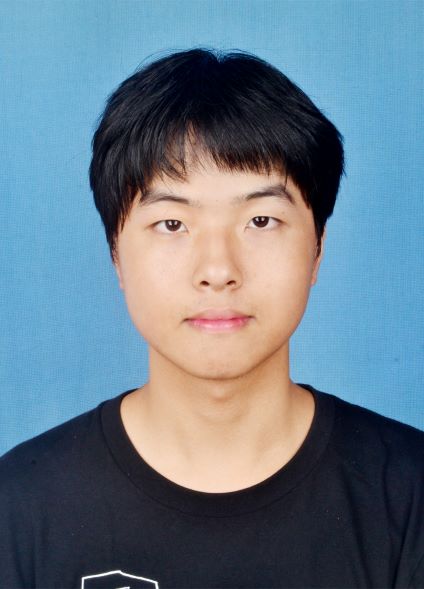}}]{Ao Li} received the B.E. degree in Artificial Intelligence from Xidian University, Xi'an, China, in 2024. He is currently pursuing the Ph.D. degree with the School of Artificial Intelligence, Xidian University, Xi'an, China. His research interests include low-level vision, computer vision and deep learning.
	\end{IEEEbiography}
	
	\begin{IEEEbiography}[{\includegraphics[width=1in,height=1.25in,clip,keepaspectratio]{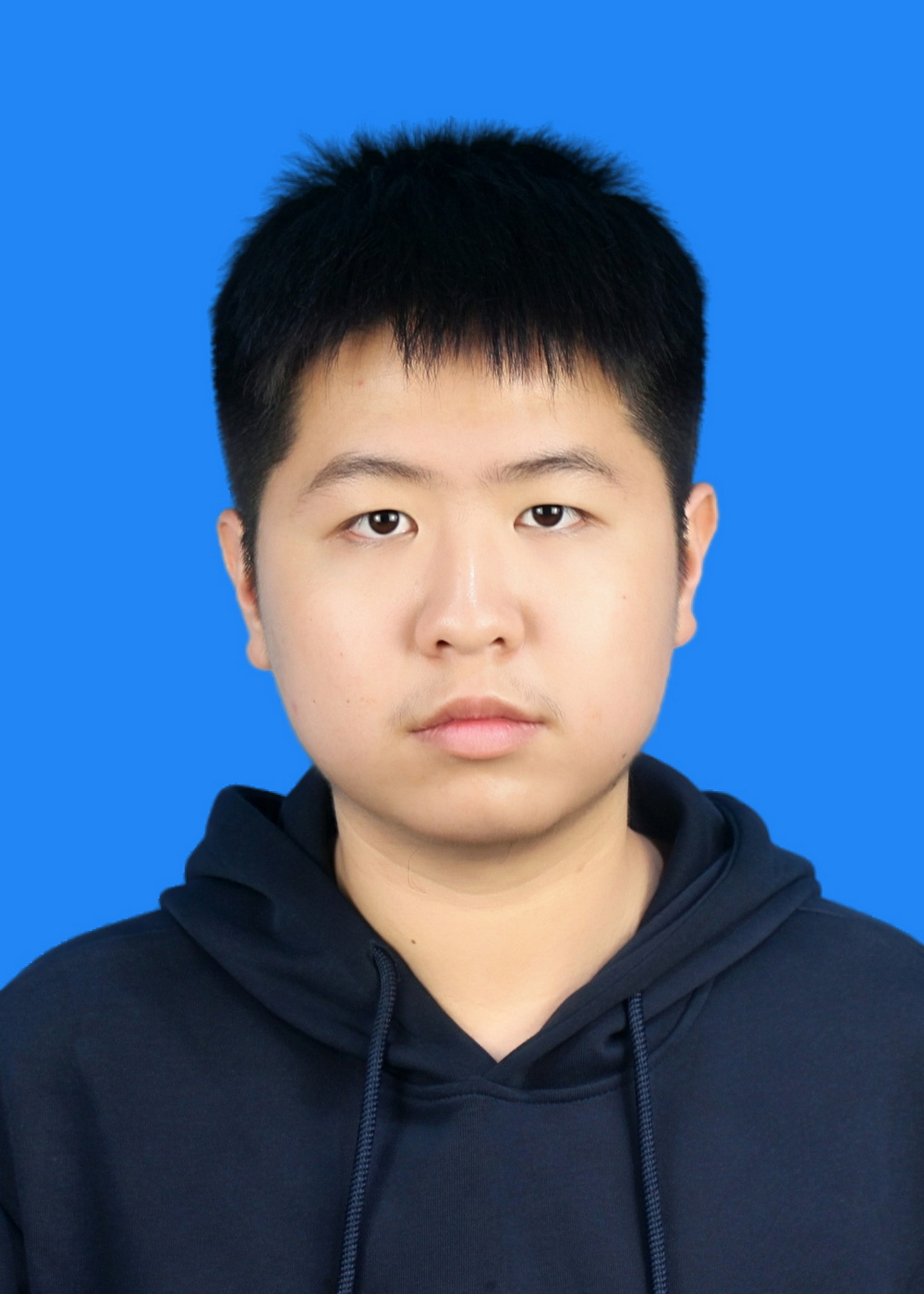}}]{Jiawei Sun} is currently pursuing the B.S. degree with the School of Artificial Intelligence, Xidian University, Xi'an, China. He is interested in computer vision and deep learning.
	\end{IEEEbiography}
	
	\begin{IEEEbiography}[{\includegraphics[width=1in,height=1.25in,clip,keepaspectratio]{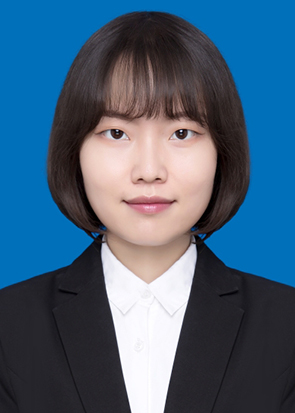}}]{Le Dong}
		received the B.S. degree in photogrammetry and remote sensing from Wuhan University, Wuhan, China, in 2016, and the Ph.D. degree in signal and information processing from the University of Chinese Academy of Sciences, Beijing, China, in 2021. She is currently an Assistant Professor with the School of Artificial Intelligence, Xidian University. Her research interests include hyperspectral image analysis, pattern recognition, and machine learning.
	\end{IEEEbiography}
	
	\begin{IEEEbiography}[{\includegraphics[width=1in,height=1.25in,clip,keepaspectratio]{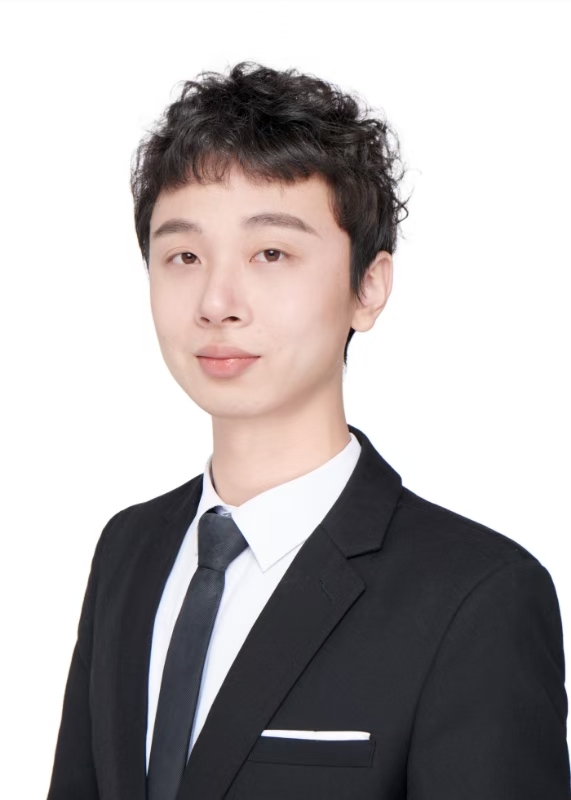}}]{Zhenyu Wang} received the Ph.D. degree from the School of Artificial Intelligence, Xidian University, Xi'an, China, in 2023, where he received the B.S. degree from the School of Microelectronics. He is currently the postdoctor in Hangzhou Institute of Technology, Xidian University. His research interests include deep
		learning, computer vision, neural networks, and model compression.
	\end{IEEEbiography}
	
	\begin{IEEEbiography}[{\includegraphics[width=1in,height=1.25in,clip,keepaspectratio]{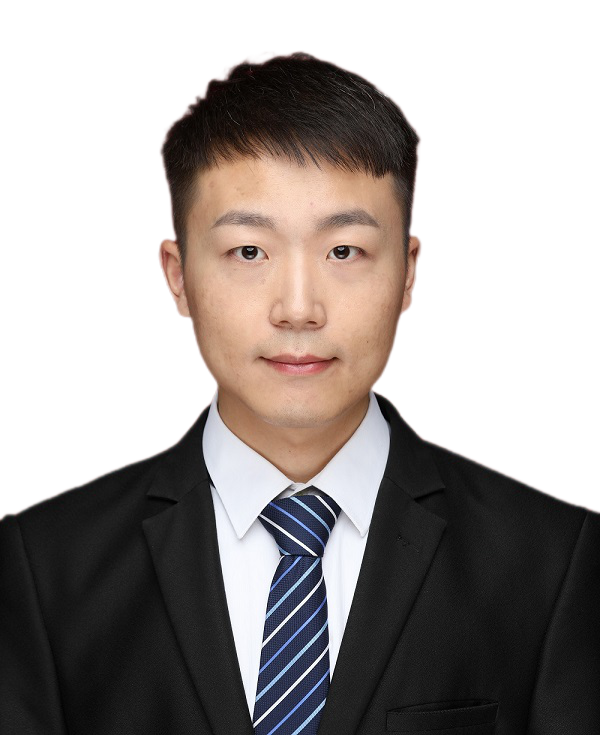}}]{Mingtao Feng}
		received the Ph.D. degree in circuit and systems from the College of Electrical and Information Engineering, Hunan University, Changsha, China, in 2019. He was a Visiting Ph.D. Student with the School of Computer Science and Software Engineering, The University of Western Australia, Perth, WA, Australia, from 2016 to 2018. He is currently an Associate Professor with the School of Artificial Intelligence, Xidian University. His research interests include 3D computer vision and point cloud understanding.
	\end{IEEEbiography}

	\begin{IEEEbiography}[{\includegraphics[width=1in,height=1.25in,clip,keepaspectratio]{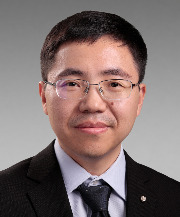}}]{Weisheng Dong}(Member, IEEE)
		received the B.S. degree in electronic engineering from the Huazhong University of Science and Technology, Wuhan, China, in 2004, and the Ph.D. degree in circuits and system from Xidian University, Xi'an, China, in 2010. He was a Visiting Student with Microsoft Research Asia, Beijing, China, in 2006. From 2009 to 2010, he was a Research Assistant with the Department of Computing, Hong Kong Polytechnic University, Hong Kong. In 2010, he joined the School of Electronic Engineering, Xidian University, as a Lecturer, where he has been a Professor since 2016. His research interests include inverse problems in image processing, sparse signal representation, and image compression. He was a recipient of the Best Paper Award at the SPIE Visual Communication and Image Processing (VCIP) in 2010.  He has served as an associate editor of IEEE Transactions on Image Processing and is currently an associate editor of SIAM Journal of Imaging Sciences.
	\end{IEEEbiography}

\begin{IEEEbiography}[{\includegraphics[width=1in,height=1.25in,clip,keepaspectratio]{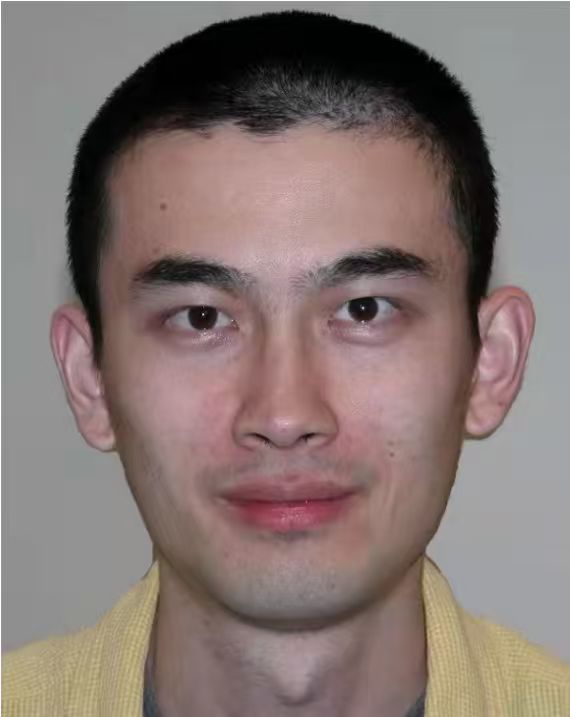}}]{Xin Li}(Fellow, IEEE)
	received the B.S. degree (Hons.) in electronic engineering and information science from the University of Science and Technology of China, Hefei, in 1996, and the Ph.D. degree in electrical engineering from Princeton University, Princeton, NJ, USA, in 2000. He was a member of Technical Staff with the Sharp Laboratories of America, Camas, WA, USA, from August 2000 to December 2002. Since January 2003, he has been a Faculty Member with the Lane Department of Computer Science and Electrical Engineering. In 2023, he joined the Department of Computer Science, University at Albany. He was elected as a fellow of IEEE in 2017, for his contributions to image interpolation, restoration, and compression.
\end{IEEEbiography}

\end{document}